\documentclass[conference]{IEEEtran}
\IEEEoverridecommandlockouts
\usepackage{cite}
\usepackage{amsmath,amssymb,amsfonts}
\usepackage{algorithmic}
\usepackage{float}
\usepackage{graphicx}
\usepackage{hyperref}
\usepackage{stfloats}
\usepackage{textcomp}
\usepackage{xcolor}
\def\BibTeX{{\rm B\kern-.05em{\sc i\kern-.025em b}\kern-.08em
    T\kern-.1667em\lower.7ex\hbox{E}\kern-.125emX}}
\begin{document}

\title{Trustworthy Chronic Disease Risk Prediction For Self-Directed Preventive Care via Medical Literature Validation}

\author{
\IEEEauthorblockN{Minh Le}
\IEEEauthorblockA{
\textit{Georgia Institute of Technology}\\
Atlanta, GA, United States \\
minhle@gatech.edu}

\and

\IEEEauthorblockN{Khoi Ton}
\IEEEauthorblockA{
\textit{University of South Florida}\\
Tampa, FL, United States \\
khoiminhtonthat@usf.edu}
}

\maketitle

\begin{abstract}
Chronic diseases are long-term, manageable, yet typically incurable conditions, highlighting the need for effective preventive strategies. Machine learning has been widely used to assess individual risk for chronic diseases. However, many models rely on medical test data (e.g. blood results, glucose levels), which limits their utility for proactive self-assessment. Additionally, to gain public trust, machine learning models should be explainable and transparent. Although some research on self-assessment machine learning models includes explainability, their explanations are not validated against established medical literature, reducing confidence in their reliability.

To address these issues, we develop deep learning models that predict the risk of developing 13 chronic diseases using only personal and lifestyle factors, enabling accessible, self-directed preventive care. Importantly, we use SHAP-based explainability to identify the most influential model features and validate them against established medical literature.

Our results show a strong alignment between the models' most influential features and established medical literature, reinforcing the models' trustworthiness. Critically, we find that this observation holds across 13 distinct diseases, indicating that this machine learning approach can be broadly trusted for chronic disease prediction.

This work lays the foundation for developing trustworthy machine learning tools for self-directed preventive care. Future research can explore other approaches for models' trustworthiness and discuss how the models can be used ethically and responsibly.
\end{abstract}

\begin{IEEEkeywords}
trustworthy, chronic disease, risk prediction, self-directed preventive care, literature validation, machine learning, deep learningdeep learning, explainability, preventive care, chronic disease
\end{IEEEkeywords}

\section{Introduction}

Chronic diseases are long-term, manageable, yet typically incurable conditions \cite{b12}, highlighting the need for effective preventive strategies. Machine learning has been widely used to assess individual risk for chronic diseases. However, many models rely on medical test data (e.g. blood results, glucose levels), which limits their utility for proactive self-assessment. Additionally, to gain public trust, machine learning models should be explainable and transparent. Although some research on self-assessment machine learning models includes explainability, their explanations are not validated against established medical literature, reducing confidence in their reliability.

\vspace{1mm}
This paper introduces trustworthy deep learning models that predict the risk of developing 13 chronic diseases using only non-clinical personal and lifestyle factors. Our approach is novel in that:

\begin{itemize}
    \item We identify the most influential features for the models and \textbf{validate them against established medical literature}, supporting the trustworthiness of the models.
    \item We demonstrate that this validation \textbf{holds across all 13 chronic diseases considered}, indicating the general trustworthiness of our approach for chronic disease prediction broadly.
\end{itemize}

\subsection{Background}

Chronic diseases include diabetes, high cholesterol, high blood pressure, depressive disorder, and stroke, among many others. Chronic diseases present a significant global health challenge, with estimates indicating that 76.4\% of adults in the United States have one or more chronic conditions \cite{b13}, calling for more effective preventive measures. 

\vspace{1mm}
Machine learning methods have been widely implemented to assess the risk of developing chronic diseases for preventive care. This risk assessment predicts an individual’s susceptibility to certain chronic diseases, which can serve as a preemptive warning and encourage lifestyle changes to avoid health repercussions.

\subsection{Research Gap}

\vspace{1mm}
\textbf{Usage of medical tests.} Most research uses medical tests as part of the input to their machine learning models because these tests provide valuable information for predicting chronic disease risk. For example, Debal \& Sitote \cite{b1} used blood urea nitrogen and serum creatinine for kidney disease prediction. Wang et al. \cite{b2} utilized blood test results and biochemical indicators to predict cardiovascular diseases. Kumar et al. \cite{b3} used plasma glucose concentration for diabetes prediction.

\vspace{1mm}
Although these models are accurate and useful, they are less suited for self-directed preventive care. This is because the need for medical testing makes these models too cumbersome to use for self-assessment. To support self-directed care, models should only rely on non-medical features, such as personal and lifestyle information.

\vspace{1mm}
\textbf{Trustworthy prediction.} Because the outputs of self-directed care tools often lack professional validation, they may raise doubt among users and reduce user trust \cite{b16}. Thus, developing trustworthy tools for personal usage in healthcare is especially important and useful for self-directed care.

\vspace{1mm}
To enhance trust in machine learning models, they should be explainable to enhance their transparency \cite{b17}. Among models suitable for self-directed care, Allani \cite{b14} used explainability to provide personalized insights and recommendations for diabetes prediction, whereas Akter et al. \cite{b15} used explainability to identify the most influential features of a machine learning model for heart disease prediction.

\vspace{1mm}
However, laypeople may still distrust these explanations, as they too are unverified by medical professionals. To the best of our knowledge, no prior work on self-directed preventive care validates model explanations against medical literature to establish trustworthiness.

\subsection{Contributions and Comparison with the State of the Art}

In this paper, we propose that the behavior of machine learning models, as observed through explanations, should be validated against the literature. Such validation, if successful, will provide strong evidence to ensure and communicate the models' trustworthiness to the lay users.

\vspace{1mm}
The contributions of this paper include:
\begin{itemize}
    \item The development of deep learning models to predict the risk of developing 13 different chronic diseases using only non-medical input.
    \item The validation of the most influential features identified by the models against established medical literature (using the SHAP explainability method), thus supporting the trustworthiness of the models.
    \item The demonstration that this validation holds across all 13 chronic diseases considered, suggesting that this machine learning approach can be broadly trusted for chronic disease prediction in general.
\end{itemize}

To the best of our knowledge, this is the first paper to develop and validate a machine learning model for self-directed preventive care from a trustworthiness perspective. While other state-of-the-art research does use explainability methods for their models, our paper is the first to validate the models' explanation against current medical literature to demonstrate the models' trustworthiness.

\section{Dataset and Data Processing}

\subsection{Dataset Description}

We use the 2023 Behavioral Risk Factor Surveillance System (BRFSS) dataset \cite{b18} to train our machine learning models.

\vspace{1mm}
The BRFSS is the most comprehensive annual telephone-based health survey system. It provides data from participants in all 50 U.S. states, the District of Columbia, and other U.S. territories. With over 400,000 interviews, the dataset is a rich source for healthcare machine learning research.

\vspace{1mm}
The survey consists of three main components. First, the core component consists of questions administered by all states. The questions cover demographics, health conditions, and lifestyle habits. Next, the optional modules contain questions on specific issues that each state may choose to include or exclude. Finally, there are additional state-specific questions as decided by each state. In this study, we only use the core component, as the other modules have substantial missing data due to optional state participation. We also exclude state-specific questions, as data for such questions are limited to the specific state.

\vspace{1mm}
We choose this dataset for three main reasons. (1) It is a large and comprehensive dataset with over 433,000 entries, allowing more data to be used to train the ML model. (2) The dataset contains many useful and relevant fields for risk prediction, as it is the result of a health survey. (3) The dataset is derived from a survey without medical test questions, which is ideal for our study that focuses on self-directed care.

\subsection{Choice of Features}

This section discusses the features that are used by our machine learning model to predict the risk of developing chronic diseases.

\vspace{1mm}
We consider most health-related fields from the core component of the survey and discard fields that contain region-specific information. For example, we exclude military service status: with different countries having different military organizations and extent of involvement, it is difficult to generalize this question for an international audience, and is therefore excluded from the study.

\vspace{1mm}
We use 38 input fields, summarized and categorized in Table \ref{table:ml_input}. These inputs are used to predict the risk of developing 13 chronic diseases present in the dataset listed in Table \ref{table:ml_diseases}.

\begin{table}[h]
    \centering
    \caption{Categories of Fields Used as Model Input}
    \begin{tabular}{|p{2.5cm}|p{5.5cm}|}
        \hline
        \textbf{Category} & \textbf{Inputs} \\
        \hline
        General & General, physical and mental health; poor health preventing usual activities.\\
        \hline
        Healthcare access & Personal doctor, cannot afford healthcare, last routine checkup.\\
        \hline
        Exercising & Most and second most often exercise type, frequency, and duration of session; strength training.\\
        \hline
        Personal information & Sex, marital status, education, own or rent home, employment, weight, height.\\
        \hline
        Disability & Deaf, blind; difficulty making decisions, climbing stairs, dressing, doing errands.\\
        \hline
        Smoking habit & Amount and frequency of smoking; use of chewing tobacco/snuff/snus and e-cigarettes.\\
        \hline
        Drinking habit & Drinking and heavy drinking day per month; drinks per drinking day.\\
        \hline
        COVID-19 & Contracted COVID-19, long-lasting symptoms, reduced ability to carry out daily activities.\\
        \hline
    \end{tabular}
    \label{table:ml_input}
\end{table}

\begin{table}
    \centering
    \caption{List of Chronic Diseases Considered and Their Labels}
    \begin{tabular}{|p{1.5cm}|p{6.5cm}|}
        \hline
        \textbf{Label} & \textbf{Disease} \\
        \hline
        BPHIGH6 & High blood pressure\\
        \hline
        TOLDHI3 & High cholesterol\\
        \hline
        CVDINFR4 & Heart attack\\
        \hline
        CVDCRHD4 & Angina or coronary heart disease\\
        \hline
        CVDSTRK3 & Stroke\\
        \hline
        ASTHMA3 & Asthma\\
        \hline
        CHCSCNC1 & Skin cancer that is not melanoma\\
        \hline
        CHCOCNC1 & Melanoma or any other types of cancer\\
        \hline
        CHCCOPD3 & Chronic obstructive pulmonary disease, emphysema, or chronic bronchitis\\
        \hline
        ADDEPEV3 & Depressive disorder\\
        \hline
        CHCKDNY2 & Kidney disease (not including kidney stones, bladder infection or incontinence)\\
        \hline
        HAVARTH4 & Arthritis, rheumatoid arthritis, gout, lupus, or fibromyalgia\\
        \hline
        DIABETE4 & Diabetes\\
        \hline 
    \end{tabular}
    \label{table:ml_diseases}
\end{table}

\subsection{Data Cleaning and Processing}

To ensure data quality for accurate machine learning model training, we perform data cleaning and processing as discussed below.

\vspace{1mm}
We remove all data points with missing information in any of the fields listed in Table \ref{table:ml_input}. It should be noted that, for this dataset, an empty field does not always indicate missing information. For example, if a surveyee responds that they did no exercises, they will not be asked further questions on exercising. Here, we still consider that information exists on those further questions and assume a response of ``no exercising'' (even though the field might be empty). Conversely, some non-empty fields represent non-responses (e.g. a value 99 might indicate the refusal to answer), which we treat as missing information (even though the field is not empty). We find that there are a total of 154,475 data points without missing information, making it still a sufficiently large dataset that is suitable for model training.

\vspace{1mm}
There are some values used to indicate the answer \textit{``None''} rather than a typical categorical or numerical response, in which case we map these to more appropriate values. For example, in a question on physical health, the value 88 is used to indicate that the surveyee has zero days in the past month in which their physical health was not good. Here, we reassign the value 0 to indicate ``zero days'' of poor physical health.

\vspace{1mm}
Some questions have multiple different units being recorded. For example, a question accepts both ``times per week'' and ``times per month.'' Other questions record responses with both hours and minutes. We convert all responses to a consistent unit to maintain a linear scale.

\section{Machine Learning Methodology}

\subsection{Machine Learning Problem and Model Architecture}

Given an input array of size 38 representing the personal and lifestyle data of an individual (Table \ref{table:ml_input}), we use a deep learning model to predict the risk (between 0 and 1) of developing each of the 13 chronic diseases (Table \ref{table:ml_diseases}). This risk is not a true probability but a relative score: values above 0.5 suggest above-average susceptibility.

\begin{figure}
    \centering
    \includegraphics[width=\linewidth]{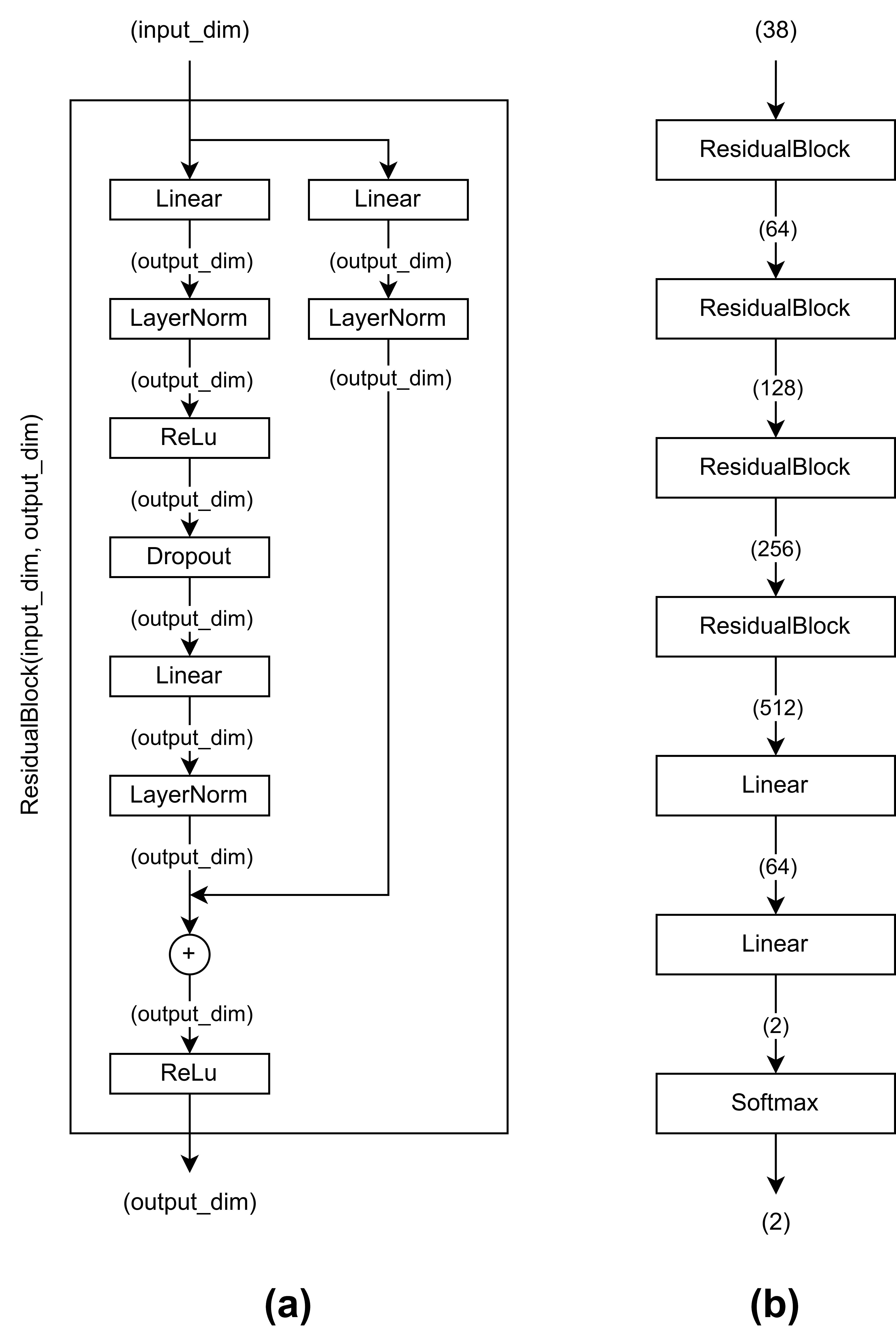}
    \caption{\textbf{(a)} Architecture of the residual block, forming part of the machine learning model; \textbf{(b)} Architecture of the entire machine learning model.}
    \label{fig:model}
\end{figure}

\vspace{1mm}
We choose a ResNet architecture for our deep learning model. The model makes use of residual blocks, whose detailed architecture is shown in Figure \ref{fig:model}a, to address the issue of vanishing gradients in deep learning models. 

\vspace{1mm}
The full deep learning model architecture is shown in Figure \ref{fig:model}b. The model takes the 38 input features and passes them through residual blocks, before passing the output of those layers through another two linear layers. A softmax layer ensures they sum to 1; the second value represents the risk score.

\vspace{1mm}
We encounter a class imbalance issue, where for all diseases, the percentage of the survey population without the disease takes the majority, and by a large ratio for some diseases. To mitigate this, we use a weighted cross-entropy loss during model training, where the weight for each class is inversely proportional to its frequency in the dataset.

\vspace{1mm}
Beyond tracking the train and test loss, we also evaluate the model as a classifier to compute accuracy and recall, thus having more meaningful and interpretable metrics of the model. We use the two final output values of the model, which correspond to the two classes of not having and having the disease, respectively. Then, we take the class with the higher representative value to be the model classification prediction. By calculating both the accuracy and recall of the model for this classification task, we ensure that the model performs well despite the class imbalance.

\subsection{Explainability for Verification with Medical Literature}

In order to demonstrate the trustworthiness of our machine learning models, we use an explainability tool called SHAP (Shapley Additive exPlanations) to identify the three most influential factors of the model for each disease. We aim to verify using the medical literature that these factors have strong correlations with each corresponding disease.

\vspace{1mm}
We use SHAP to measure the extent to which each input influences the model's prediction. SHAP does this by calculating how much each input factor changes the final prediction, via measuring the expected difference when that factor is varied, using other data points as background references. Specifically, we use SHAP's Kernel Explainer that uses a linear regression to calculate this expected difference. This results in a SHAP value for each factor of a specific input. We can deduce the explainability of the entire machine learning model by taking the average SHAP values of each factor across all data points in a dataset.

\vspace{1mm}
Due to computational limits, we explain predictions for 500 randomly selected points and sum up their SHAP values for overall model explanations. The background data used for this explanation is also reduced from the entire dataset to 100 representative data points via k-means clustering.

\section{Results}

\subsection{Machine Learning Model Training}

We train machine learning models using data batches of 32, with a weighted cross-entropy loss inversely proportional to class frequency to address the imbalance. We use the Adam optimizer, with an initial learning rate of 0.001 (halved if train loss stagnates for 3 epochs), to train the model for 50 epochs. The model with the lowest test loss is selected.

\vspace{1mm}
Table \ref{table:ml_result} shows the training results of the models for all 13 chronic diseases. For all diseases, the test loss is only negligibly larger than the train loss. This shows that our models do not overfit, and they can generalize well for new data. In using the models for the classification task, most models have accuracy and recall values between 65\% - 75\%. This is comparable to other similar research that uses datasets without medical information, such as Budhathoki et al. \cite{b6}. Thus, the performance of the models isare comparable to state- of- the- art research with similar approaches, indicating that our models are sufficiently accurate for self-directed preventive care.

\vspace{1mm}
\begin{table}[b]
    \centering
    \caption{Machine Learning Model Training Results}
    \begin{tabular}{|c|c|c|c|c|}
        \hline
        \textbf{Disease} & \textbf{Train loss} & \textbf{Test loss} & \textbf{Accuracy} & \textbf{Recall}\\
        \hline 
        BPHIGH6 & 0.5802 & 0.5810 & 69.60\% & 72.50\% \\
        \hline
        TOLDHI3 & 0.6384 & 0.6439 & 61.68\% & 71.34\% \\
        \hline
        CVDINFR4 & 0.5381 & 0.5387 & 74.32\% & 70.73\% \\
        \hline
        CVDCRHD4 & 0.5295 & 0.5379 & 71.21\% & 76.32\% \\
        \hline
        CVDSTRK3 & 0.5515 & 0.5604 & 78.02\% & 62.62\% \\
        \hline
        ASTHMA3 & 0.6509 & 0.6515 & 66.80\% & 53.50\% \\
        \hline
        CHCSCNC1 & 0.6024 & 0.6066 & 66.94\% & 67.73\% \\
        \hline
        CHCOCNC1 & 0.6185 & 0.6191 & 63.99\% & 68.59\% \\
        \hline
        CHCCOPD3 & 0.4893 & 0.4896 & 77.90\% & 76.17\% \\
        \hline
        ADDEPEV3 & 0.5078 & 0.5151 & 76.54\% & 72.68\% \\
        \hline
        CHCKDNY2 & 0.5544 & 0.5675 & 74.72\% & 68.66\% \\
        \hline
        HAVARTH4 & 0.5642 & 0.5660 & 70.64\% & 70.23\% \\
        \hline
        DIABETE4 & 0.5390 & 0.5408 & 72.24\% & 73.38\% \\
        \hline
    \end{tabular}
    \label{table:ml_result}
\end{table}

\subsection{Explainability of Machine Learning Models}

We use SHAP to explain of our machine learning models, in order to identify the most influential factor for the development of each chronic disease. Table \ref{tab:global_explain} shows the three most influential factors for each chronic disease. Notably, general health, physical health, and days with poor health are excluded, because they are more likely to be consequences, rather than causes, of chronic diseases.

\begin{table*}
    \centering
    \caption{Factors with the most influence on chronic diseases}
    \begin{tabular}{|c|c|c|c|}
        \hline
        \textbf{Disease} & \textbf{Top predictor} & \textbf{2nd top predictor} & \textbf{3rd top predictor} \\
        \hline
        High blood pressure & Weight & Employment & Marital status \\
        \hline
        High cholesterol & Employment & Weight & Routine checkup \\
        \hline
        Heart attack & Employment & Sex & Smoking \\
        \hline
        Coronary heart disease & Employment & Sex & Smoking \\
        \hline
        Stroke & Employment & Exercising & Difficulty walking \\
        \hline
        Asthma & Weight & Mental health & Sex \\
        \hline
        Non-melanoma skin cancer & Employment & Days consuming alcohol & Education \\
        \hline
        Other cancers & Employment & Education & Smoking \\
        \hline
        Chronic bronchitis & Smoking & Employment & Exercising duration \\
        \hline
        Depressive disorder & Mental health & Sex & Weight \\
        \hline
        Kidney disease & Employment & Days consuming alcohol & Personal doctor \\
        \hline
        Arthritis & Employment & Marital status & Sex \\
        \hline
        Diabetes & Weight & Employment & Days consuming alcohol \\
        \hline
    \end{tabular}
    \label{tab:global_explain}
\end{table*}

\vspace{1mm}
Before validating these factors with the medical literature, we perform some further data analysis to make more sense of these factors.

\vspace{1mm}
\subsubsection{Employment}
This factor belongs in the top 3 most influential factors for all but 2 diseases (namely, asthma and depressive disorder) and is the top predictor factor for 8 out of 13 diseases.

\vspace{1mm}
While initially surprising, we find that this factor actually serves as an indicator of age and the ability to work in this context. This is due to the answer choices that include \textit{A student}, \textit{Retired}, and \textit{Unable to work}. Since the inability to work is more likely to be a consequence rather than a cause of chronic disease, we focus on analyzing this field as an indicator of age instead.

\vspace{1mm}
Table \ref{tab:employment} reveals that for 11 out of 13 chronic diseases (excluding asthma and depressive disorder) where employment is a top-three influential factor, students consistently show a substantially lower percentage of disease prevalence, while retirees show a substantially higher percentage. This indicates that \textit{employment serves as an indicator that old age increases the risk of developing most diseases}, and this interpretation is consistent with medical literature validation. 

\begin{table}
    \centering
    \caption{Percentage of Students and Retirees with Chronic Diseases}
    \begin{tabular}{|c|c|c|c|}
        \hline
        \textbf{Disease} & \textbf{Student} & \textbf{Retiree} & \textbf{Others} \\
        \hline
        High blood pressure & 14.09 & 60.67 & 32.59 \\
        \hline
        High cholesterol & 15.84 & 55.46 & 34.58 \\
        \hline
        Heart attack & 0.84 & 9.54 & 2.65 \\
        \hline
        Coronary heart disease & 0.76 & 10.87 & 2.60 \\
        \hline
        Stroke & 0.56 & 6.95 & 1.99 \\
        \hline
        Asthma & 19.71 & 13.16 & 14.57 \\
        \hline
        Non-melanoma skin cancer & 0.68 & 17.09 & 5.09 \\
        \hline
        Other cancers & 1.51 & 22.02 & 7.35 \\
        \hline
        Chronic bronchitis & 2.35 & 12.13 & 3.95 \\
        \hline
        Depressive disorder & 29.86 & 17.24 & 21.03 \\
        \hline
        Kidney disease & 0.84 & 8.77 & 2.43 \\
        \hline
        Arthritis & 7.84 & 52.87 & 23.59 \\
        \hline
        Diabetes & 2.07 & 21.78 & 10.04 \\
        \hline
    \end{tabular}
    \label{tab:employment}
\end{table}

\subsubsection{Sex}
Five of the 13 chronic diseases have sex in the three most influential factors: heart attack, coronary heart disease, asthma, depressive disorder, and arthritis.

\vspace{1mm}
Table \ref{tab:sex} shows that males are more susceptible to heart attack and coronary heart disease, while females are more at risk forto asthma, depressive disorder, and arthritis.

\vspace{1mm}
\subsubsection{Marital status}
High blood pressure and arthritis have marital status as the third and second most influential factors, respectively. In a similar way to the employment factor, we find that this is also an indicator of age.

\vspace{1mm}
Divorcees and widows show higher rates of high blood pressure and arthritis, with widows more impacted than divorcees. This suggests age ias a factor, as divorcees are typically older than married/unmarried individuals, and widows are generally older than divorcees. Thus, \textit{marital status serves as an age indicator}, similar to employment.

\subsection{Validation with Medical Literature}

We find that the medical literature agrees that all of the most influential factors of the machine learning models, as presented in Table \ref{tab:global_explain}, are indeed significant causes of their respective chronic diseases. 

\vspace{1mm}
\subsubsection{High blood pressure}
Seravalle \& Grassi \cite{b19} noted that obesity changes the body's ``hormonal, inflammatory, and endothelial level,'' contributing to the development of high blood pressure. In addition, according to Cheng et al. \cite{b21}, the likelihood of elevated blood pressure increases continuously from the ages of 35 to 79. This explains both employment and marital status being top predictors of our model, as they are representative of age.

\vspace{1mm}
\subsubsection{High cholesterol}
Bertolotti et al. \cite{b22} found a steady increase in total blood cholesterol after age 20, linking it to age. Miettinen \cite{b23} demonstrated obesity's connection to higher cholesterol production. Additionally, Liss et al. \cite{b24} showed routine health checkups moderately improve the management of chronic disease risk factors, including cholesterol.

\begin{table}
    \centering
    \caption{Percentage of Each Gender with Selected Chronic Diseases}
    \begin{tabular}{|c|c|c|}
        \hline
        \textbf{Disease} & \textbf{Male} & \textbf{Female} \\
        \hline
        Heart attack & 7.79 & 4.17 \\
        \hline
        Coronary heart disease & 8.10 & 4.71 \\
        \hline
        Asthma & 11.92 & 17.64 \\
        \hline
        Depressive disorder & 15.57 & 26.72 \\
        \hline
        Arthritis & 31.09 & 40.56 \\
        \hline
    \end{tabular}
    \label{tab:sex}
\end{table}

\begin{table}
    \centering
    \caption{Percentage of those with Selected Chronic Diseases by Marital Status}
    \begin{tabular}{|c|c|c|c|}
        \hline
        \textbf{Disease} & \textbf{Divorced} & \textbf{Widowed} & \textbf{Others} \\
        \hline
        High blood pressure & 50.41 & 63.30 & 40.08 \\
        \hline
        Arthritis & 44.88 & 56.51 & 31.83 \\
        \hline
    \end{tabular}
    \label{tab:marital_status}
\end{table}

\vspace{1mm}
\subsubsection{Heart attack and Coronary heart disease}
Because both of these diseases are heart diseases and have the same three most influential factors, we will treat them together as heart disease in general. Stern et al. \cite{b25} highlighted that aging leads to fat and cholesterol accumulation in artery walls, raising heart disease risk. Kannel's research \cite{b50} showed that the incidence of coronary heart disease is two to threefold higher in those aged 35-64 compared to 65-94. The research also showed that smoking is a known promoter of hazardous atherosclerosis.

\vspace{1mm}
\subsubsection{Stroke}
Kelly-Hayes \cite{b26} demonstrated stroke risk doubles each decade after 45, with over 70\% of strokes occurring in those 65 and above. Kelly-Hayes also showed that physical activity reduces stroke risk; moderately active adults have a 20\% lower risk, and highly active adults have a 27\% lower risk. Thus, lack of movement is a risk factor.

\vspace{1mm}
\subsubsection{Asthma}
Boulet \cite{b27} found that obesity increases asthma risk. Liu et al. \cite{b29} showed that asthma and mental health disorders have a bidirectional link, possibly due to shared environmental factors, asthma-induced anxiety, and increased inflammatory cytokines. Yung et al. \cite{b30} found that starting around puberty, females have an increased risk of contracting asthma.

\vspace{1mm}
\subsubsection{Non-melanoma skin cancer (NSMC)}
NMSC is most common in the elderly, as shown by Byfield et al. \cite{b31}. Alcohol's metabolite, acetaldehyde, is carcinogenic by damaging DNA (Darvin et al. \cite{b32}), confirming alcohol's link to NMSC (Hezaveh et al. \cite{b33}). Counterintuitively, higher socioeconomic development increases NMSC risk (Hu et al. \cite{b34}), possibly due to more tanning or better detection, suggesting a strong link to education level.

\vspace{1mm}
\subsubsection{Other cancers}
DePinho's research \cite{b35} indicated that age is the leading carcinogen, with cancer incidence rising exponentially from age 40 to 80 years old. Coughlin \cite{b37} determined that a lower socio-economic status iwas associated with an increased risk of colorectal cancer, which also indicates a link to education level. Sasco et al. \cite{b36} found that 86.8\% of lung cancer deaths stem from tobacco smoking, which also causes other cancers like colorectal cancer and leukemia.
 
\vspace{1mm}
\subsubsection{Chronic bronchitis (CB)}
Forey et al. \cite{b38} found that ever-smokers have a 2.69 times higher relative risk of CB, and current smokers have a 3.41 times higher risk. Ferré et al. \cite{b39} confirmed the age-CB link and smoking's impact. In addition, Wan et al. \cite{b51} showed that CB patients with chronic bronchitis who haves greater exercise tolerance haves milder symptoms.
 
\vspace{1mm}
\subsubsection{Depressive disorder}
Mental health is the leading factor for depressive disorder risk. Piccinelli \& Wilkinson \cite{b41} showed higher female risk due to e.g. adverse childhood experiences, prior youth depression/anxiety, and sociocultural roles. Furthermore, Marx et al. \cite{b40} noted that behavioral factors like obesity can act as psychological contributors to depression. 

\vspace{1mm}
\subsubsection{Kidney diseases}
Levey et al. \cite{b42} identified age as a major contributor to kidney disease in the elderly, affecting over half of adults over 70 chronically. Shankar et al. \cite{b52} found a relationship between heavy alcohol consumption and chronic kidney disease risk. ``Personal doctor'' as a factor indicates socioeconomic status, as lower income correlates with fewer healthcare providers. Krop et al. \cite{b43} reported that individuals with incomes less than \$16,000 had a 2.4-fold increased risk of chronic kidney disease compared to those with incomes more than \$35,000.

\vspace{1mm}
\subsubsection{Arthritis}
Boots et al. \cite{b44} found that aging, through immunosenescence, profoundly alters the immune system, impacting RA onset and presentation. Favalli et al. \cite{b45} showed RA is more common and severe in women, with frequent female comorbidities like fibromyalgia, depression, and osteoporosis potentially affecting treatment choices and outcomes.

\vspace{1mm}
\subsubsection{Diabetes}
Sherr and Lipman's \cite{b46} research indicated that a BMI of 25\% or higher increases diabetes risk due to reduced cellular insulin responsiveness from fatty tissues. Furthermore, Boles et al. \cite{b47} found that being 45 years or older also elevates diabetes risk. van de Wiel's work on alcohol and diabetes\cite{b48} revealed that excessive alcohol consumption can lead to loss of metabolic control and negate alcohol's beneficial cardiovascular effects.

\section{Limitations and Future Work}

We discuss a few limitations of our methodology in developing trustworthy machine learning tools for self-directed preventive care.

\vspace{1mm}
Firstly, we approach trustworthiness from the explainability angle, in which we validate the model explanations against medical literature, thus enabling explainability to enhance trustworthiness. That said, there are many other approaches to developing trustworthiness, such as ensuring robustness to data shifts and to adversarial attacks \cite{b55}. Future work can explore these different approaches to develop more trustworthy models.

\vspace{1mm}
In addition, although our dataset is large and comprehensive, it is limited to residents of the United States. Future work can apply this method to datasets from other countries and explore whether the model explanations can still be validated with medical literature. This is especially relevant if there is a lack of an equally large and comprehensive dataset, resulting in lower-quality data for machine learning model training.

\vspace{1mm}
Finally, a model's alignment with medical literature doesn't guarantee trustworthiness, as it might reinforce existing chronic disease risk patterns, potentially causing over- or under-caution in health decisions. For instance, despite Table \ref{tab:sex} suggesting women are less susceptible to heart disease, Mass \& Appelman \cite{b49} contend this perception leads to ``under-recognition of heart disease'' and ``less aggressive treatment strategies'' for women. Future research could explore how self-directed tools can avoid perpetuating such biases.

\section{Conclusion}

In this work, we develop trustworthy machine learning models for self-directed preventive care for chronic diseases by validating model explanations against medical literature. We find that the validation holds across 13 distinct chronic diseases, indicating that this machine learning approach can be broadly trusted for chronic disease prediction. We believe that our work can serve as a foundation for other approaches to trustworthiness, applications in diverse countries and populations, and methods to address unintended consequences of the models.

\section*{Acknowledgment}
We would like to thank Dr. Phuong Cao, Dr. Caitlin Kelly, Dr. Anila Shree, Ms. Keiko Ang, Mr. Khanh Bui, and Ms. Katherine Bahagia for their invaluable comments on the early drafts of this paper.


\begin{thebibliography}{00}
\bibitem{b12}“About chronic diseases,” Centers for Disease Control and Prevention, https://www.cdc.gov/chronic-disease/about/ (accessed Jun. 18, 2025). 
\bibitem{b13} K. B. Watson \textit{et al.}, “Trends in Multiple Chronic Conditions Among US Adults, By Life Stage, Behavioral Risk Factor Surveillance System, 2013-2023,” Preventing Chronic Disease, vol. 22, Apr. 2025, doi: https://doi.org/10.5888/pcd22.240539.
\bibitem{b1} D. A. Debal and T. M. Sitote, “Chronic kidney disease prediction using machine learning techniques,” Journal of Big Data, vol. 9, no. 1, Nov. 2022, doi: https://doi.org/10.1186/s40537-022-00657-5.
\bibitem{b2} Z. Wang \textit{et al.}, “Construction of machine learning diagnostic models for cardiovascular pan-disease based on blood routine and biochemical detection data,” Cardiovascular Diabetology, vol. 23, no. 1, Sep. 2024, doi: https://doi.org/10.1186/s12933-024-02439-0.
\bibitem{b3} N. Kumar \textit{et al.}, “Efficient Automated Disease Diagnosis Using Machine Learning Models,” Journal of Healthcare Engineering, vol. 2021, pp. 1–13, May 2021, doi: https://doi.org/10.1155/2021/9983652.
\bibitem{b16} T. P. Quinn, M. Senadeera, S. Jacobs, S. Coghlan, and V. Le, “Trust and medical ai: The challenges we face and the expertise needed to overcome them,” \textit{Journal of the American Medical Informatics Association}, vol. 28, no. 4, pp. 890–894, Dec. 2020. doi:10.1093/jamia/ocaa268 
\bibitem{b17} F. Doshi-Velez and B. Kim, “Towards a rigorous science of interpretable machine learning,” arXiv.org, https://arxiv.org/abs/1702.08608 (accessed Jun. 18, 2025). 
\bibitem{b14} U. Allani, “Interactive diabetes risk prediction using explainable machine learning: A dash-based approach with shap, lime, and comorbidity insights,” arXiv.org, https://arxiv.org/abs/2505.05683 (accessed Jun. 17, 2025). 
\bibitem{b15} S. B. Akter \textit{et al.}, \textit{Improving heart disease probability prediction sensitivity with a grow network model}, Mar. 2024. doi:10.1101/2024.02.28.24303495 
\bibitem{b18} “CDC - 2023 BRFSS survey data and Documentation,” Centers for Disease Control and Prevention, https://www.cdc.gov/brfss/annual\_data/annual\_2023.html (accessed Jun. 18, 2025).
\bibitem{b6} N. Budhathoki \textit{et al.}, “Predicting asthma using imbalanced data modeling techniques: Evidence from 2019 Michigan BRFSS data,” PLoS ONE, vol. 18, no. 12, pp. e0295427–e0295427, Dec. 2023, doi: https://doi.org/10.1371/journal.pone.0295427.
\bibitem{b19} G. Seravalle and G. Grassi, “Obesity and Hypertension,” Springer eBooks, pp. 65–79, Jan. 2024, doi: https://doi.org/10.1007/978-3-031-62491-9\_5.
\bibitem{b21} W. Cheng \textit{et al.}, “Age-related changes in the risk of high blood pressure,” Frontiers in Cardiovascular Medicine, vol. 9, Sep. 2022, doi: https://doi.org/10.3389/fcvm.2022.939103.
\bibitem{b22} M. Bertolotti \textit{et al.}, “Management of high cholesterol levels in older people,” Geriatrics \& Gerontology International, vol. 19, no. 5, pp. 375–383, Mar. 2019, doi: https://doi.org/10.1111/ggi.13647.
\bibitem{b23} T. A. Miettinen, “Cholesterol Production in Obesity,” Circulation, vol. 44, no. 5, pp. 842–850, Nov. 1971, doi: https://doi.org/10.1161/01.cir.44.5.842.
\bibitem{b24} D. T. Liss \textit{et al.}, “General Health Checks in Adult Primary Care,” JAMA, vol. 325, no. 22, p. 2294, Jun. 2021, doi: https://doi.org/10.1001/jama.2021.6524.
\bibitem{b25} S. Stern \textit{et al.}, “Aging and Diseases of the Heart,” Circulation, vol. 108, no. 14, Oct. 2003, doi: https://doi.org/10.1161/01.cir.0000086898.96021.b9.
\bibitem{b50}  W. B. Kannel, \textit{Cardiovascular Drugs and Therapy}, vol. 11, no. 1+, pp. 199–212, 1997. doi:10.1023/a:1007792820944 
\bibitem{b26} M. Kelly-Hayes, “Influence of age and health behaviors on stroke risk: Lessons from longitudinal studies,” \textit{Journal of the American Geriatrics Society}, vol. 58, no. s2, Oct. 2010. doi:10.1111/j.1532-5415.2010.02915.x 
\bibitem{b27} D. D. Sin and E. R. Sutherland, “Obesity and the lung: 4 \{middle dot\} obesity and asthma,” \textit{Thorax}, vol. 63, no. 11, pp. 1018–1023, Nov. 2008. doi:10.1136/thx.2007.086819 
\bibitem{b29}  X. Liu \textit{et al.}, “Bidirectional associations between asthma and types of mental disorders,” \textit{The Journal of Allergy and Clinical Immunology: In Practice}, vol. 11, no. 3, Mar. 2023. doi:10.1016/j.jaip.2022.11.027 
\bibitem{b30} J. A. Yung \textit{et al.}, “Hormones, sex, and asthma,” \textit{Annals of Allergy, Asthma \&amp; Immunology}, vol. 120, no. 5, pp. 488–494, May 2018. doi:10.1016/j.anai.2018.01.016 
\bibitem{b31} S. Dacosta Byfield \textit{et al.}, “Age distribution of patients with advanced non-melanoma skin cancer in the United States,” \textit{Archives of Dermatological Research}, vol. 305, no. 9, pp. 845–850, Apr. 2013. doi:10.1007/s00403-013-1357-2 
\bibitem{b32} M. E. Darvin \textit{et al.}, “Alcohol consumption decreases the protection efficiency of the antioxidant network and increases the risk of sunburn in human skin,” \textit{Skin Pharmacology and Physiology}, vol. 26, no. 1, pp. 45–51, Nov. 2012. doi:10.1159/000343908 
\bibitem{b33} E. Hezaveh \textit{et al.}, “Dietary components and the risk of non-melanoma skin cancer: A systematic review of Epidemiological Studies,” \textit{Critical Reviews in Food Science and Nutrition}, vol. 63, no. 21, pp. 5290–5305, Dec. 2021. doi:10.1080/10408398.2021.2016600 
\bibitem{b34} W. Hu \textit{et al.}, “Changing trends in the disease burden of non-melanoma skin cancer globally from 1990 to 2019 and its predicted level in 25 years,” \textit{BMC Cancer}, vol. 22, no. 1, Jul. 2022. doi:10.1186/s12885-022-09940-3 
\bibitem{b35} R. A. DePinho, “The age of cancer,” \textit{Nature}, vol. 408, no. 6809, pp. 248–254, Nov. 2000. doi:10.1038/35041694 
\bibitem{b36} A. J. Sasco \textit{et al.}, “Tobacco smoking and cancer: A brief review of recent epidemiological evidence,” \textit{Lung Cancer}, vol. 45, Aug. 2004. doi:10.1016/j.lungcan.2004.07.998 
\bibitem{b37} S. S. Coughlin, “Social determinants of colorectal cancer risk, stage, and survival: A systematic review,” \textit{International Journal of Colorectal Disease}, vol. 35, no. 6, pp. 985–995, Apr. 2020. doi:10.1007/s00384-020-03585-z 
\bibitem{b38} B. A. Forey \textit{et al.}, “Systematic review with meta-analysis of the epidemiological evidence relating smoking to COPD, chronic bronchitis and emphysema,” \textit{BMC Pulmonary Medicine}, vol. 11, no. 1, Jun. 2011. doi:10.1186/1471-2466-11-36 
\bibitem{b39} A. Ferré \textit{et al.}, “Chronic bronchitis in the general population: Influence of age, gender and socio-economic conditions,” \textit{Respiratory Medicine}, vol. 106, no. 3, pp. 467–471, Mar. 2012. doi:10.1016/j.rmed.2011.12.002 
\bibitem{b51} Q. Wan \textit{et al.}, “Association of Exercise Tolerance with Respiratory Health Outcomes in Mild-to-Moderate COPD,” Annals of the American Thoracic Society, Nov. 2024, doi: https://doi.org/10.1513/annalsats.202404-408oc.
\bibitem{b41} M. Piccinelli and G. Wilkinson, “Gender differences in depression: Critical review,” \textit{British Journal of Psychiatry}, vol. 177, no. 6, pp. 486–492, 2000. doi:10.1192/bjp.177.6.486 
\bibitem{b40} W. Marx \textit{et al.}, “Major depressive disorder,” \textit{Nature Reviews Disease Primers}, vol. 9, no. 1, Aug. 2023. doi:10.1038/s41572-023-00454-1 
\bibitem{b42} A. S. Levey \textit{et al.}, “Chronic kidney disease in older people,” \textit{JAMA}, vol. 314, no. 6, p. 557, Aug. 2015. doi:10.1001/jama.2015.6753 
\bibitem{b52} A. Shankar \textit{et al.}, “Perfluoroalkyl Chemicals and Chronic Kidney Disease in US Adults,” American Journal of Epidemiology, vol. 174, no. 8, pp. 893–900, Aug. 2011, doi: https://doi.org/10.1093/aje/kwr171.
\bibitem{b43} J. S. Krop \textit{et al.}, “A community-based study of explanatory factors for the excess risk for early renal function decline in blacks vs whites with diabetes,” \textit{Archives of Internal Medicine}, vol. 159, no. 15, p. 1777, Aug. 1999. doi:10.1001/archinte.159.15.1777
\bibitem{b44} A. M. Boots \textit{et al.}, “The influence of ageing on the development and management of rheumatoid arthritis,” \textit{Nature Reviews Rheumatology}, vol. 9, no. 10, pp. 604–613, Jun. 2013. doi:10.1038/nrrheum.2013.92 
\bibitem{b45} E. G. Favalli \textit{et al.}, “Sex and management of rheumatoid arthritis,” \textit{Clinical Reviews in Allergy \&amp; Immunology}, vol. 56, no. 3, pp. 333–345, Jan. 2018. doi:10.1007/s12016-018-8672-5 
\bibitem{b46} D. Sherr and R. D. Lipman, “Diabetes educators,” \textit{American Journal of Preventive Medicine}, vol. 44, no. 4, Apr. 2013. doi:10.1016/j.amepre.2012.12.013 
\bibitem{b47} A. Boles \textit{et al.}, “Dynamics of diabetes and obesity: Epidemiological perspective,” \textit{Biochimica et Biophysica Acta (BBA) - Molecular Basis of Disease}, vol. 1863, no. 5, pp. 1026–1036, May 2017. doi:10.1016/j.bbadis.2017.01.016 
\bibitem{b48} A. van de Wiel, “Diabetes mellitus and alcohol,” \textit{Diabetes/Metabolism Research and Reviews}, vol. 20, no. 4, pp. 263–267, Jul. 2004. doi:10.1002/dmrr.492 
\bibitem{b55} K. Rasheed \textit{et al.}, “Explainable, trustworthy, and ethical machine learning for healthcare: A survey,” Computers in Biology and Medicine, vol. 149, no. 106043, p. 106043, Oct. 2022, doi: https://doi.org/10.1016/j.compbiomed.2022.106043.
\bibitem{b49} A. H. E. M. Maas and Y. E. A. Appelman, “Gender differences in coronary heart disease,” Netherlands Heart Journal, vol. 18, no. 12, pp. 598–603, Nov. 2010, doi: https://doi.org/10.1007/s12471-010-0841-y.
\end{thebibliography}
\end{document}